# Avatar Work: Telework for Disabled People Unable to Go Outside by Using Avatar Robots "OriHime-D" and Its Verification

Kazuaki Takeuchi
Ory Laboratory Inc.
Minato Ward, Tokyo, Japan
Kanagawa Institute of Technology
Atsugi, Kanagawa, Japan
k.takeuchi@orylab.com

Yoichi Yamazaki
Kanagawa Institute of Technology
Atsugi, Kanagawa, Japan
yamazaki@he.kanagawa-it.ac.jp

Kentaro Yoshifuji
Ory Laboratory Inc.
Minato Ward, Tokyo, Japan
ory@orylab.com

## ABSTRACT

In this study, we propose a telework "avatar work" that enables people with disabilities to engage in physical works such as customer service in order to realize an inclusive society, where we can do anything if we have free mind, even though we are bedridden. In avatar work, disabled people can remotely engage in physical work by operating a proposed robot "OriHime-D" with a mouse or gaze input depending on their own disabilities. As a social implementation initiative of avatar work, we have opened a two-week limited avatar robot cafe and have evaluated remote employment by people with disabilities using OriHime-D. As the results by 10 people with disabilities, we have confirmed that the proposed avatar work leads to mental fulfillment for people with disparities, and can be designed with adaptable workload. In addition, we have confirmed that the work content of the experimental cafe is appropriate for people with a variety of disabilities seeking social participation. This study contributes to fulfillment all through life and lifetime working, and at the same time leads to a solution to the employment shortage problem.

## 1 INTRODUCTION

People feel lonely when they cannot feel connect with anyone. In order to resolve loneliness of humankind, we need to realize an inclusive society with technology, where all people can connect with people, that is, can actively participate in society in various forms throughout life.

As increasing in life expectancy, the number of people with physical disabilities or bedridden people is increasing not only because of congenital basis, but also because of acquired cause. As efforts to prevent bedridden by technology, Muscle suits [1] and HAL [2] are developed for physical support, are utilized for rehabilitation [3].

On the other hand, it is not on the table yet how to assist to lead more fulfilling life in bedridden state or how to eliminate loneliness caused by bedridden. Considering that anyone can become bedridden with age, in addition to preventions by physical support and ME-BYO care by mental and cognitive support [4][5], we need technologies that extend abilities of bedridden people, that is, technologies that enable social participation of bedridden people, which brings something to live for.

In this study, we propose a telework "avatar work" that enables people with disabilities to engage in physical works such as customer service e.g., waiter/waitress in cafe in order to realize an inclusive society, where we can do anything if we have free mind, even though we are bedridden (Figure 1). In avatar work, disabled people can remotely engage in physical work by operating a proposed robot "OriHime-D" with a mouse or gaze input depending on their own disabilities. As a social implementation initiative of avatar work, we open a two-week limited avatar robot cafe with the cooperation of 10 disabled people, and evaluate remote employment by people with disabilities using OriHime-D.

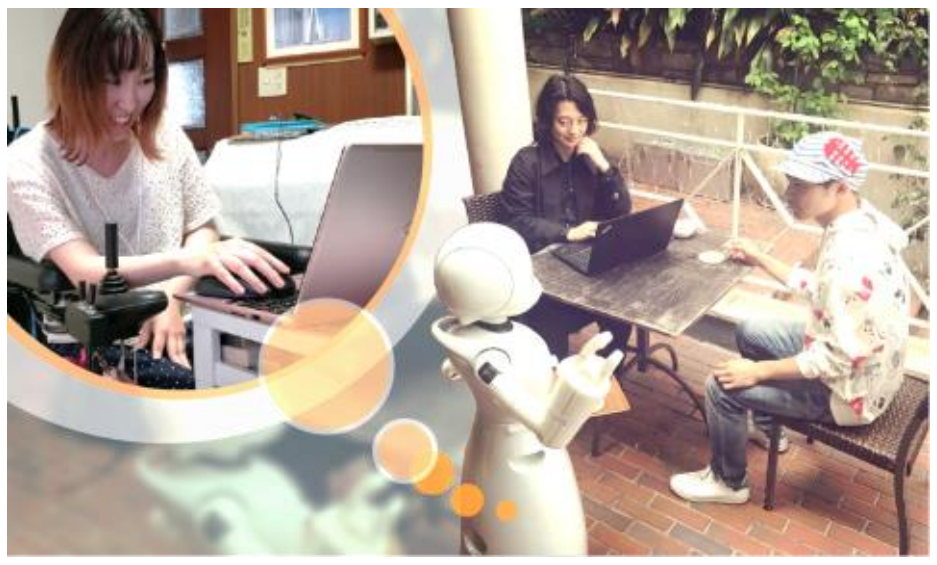

**Figure 1: Concept of Avatar work in avatar robot cafe.**

## 2 AVATAR ROBOT FOR TELEWORK BY PEOPLE WITH DISABILITIES

The robot technologies to extend the existence of uses have been realized, e.g., Model H [6] by TELEXISTENCE Inc., Beam [7] by Suitable Technologies, Inc., etc., and the Beam is used for telework.

To realize social participation by persons with disabilities e.g., bedridden persons, it is necessary to extend not only their existence beyond geographical distances like existence telework robots, but also their own abilities. This type of robot is so to speak an alter ego with new body, which updates their own body and extends their possible.

Robots as alter ego of users require (i) intuitive, simple, and easy-to-operate interface depending on his/her disability, and (ii) semi-automatic motion for acting as alter ego of the user, which can be performed by simple and easy input action. An avatar robot "OriHime" has been developed as a robot to achieve these requirements (Figure 2) [8]. User can remotely control the OriHime as his/her real avatar, that is, an alter ego with body by selecting prepared patterned motions with a mouse, a touch screen, or a gaze input. In addition, the user can communicate with not only real speech sound, but also speech synthesis. This enables communication for persons with difficulty speaking, though keeping unable to engage in physical work.

We define this type of semi-automated avatar robot that extend human abilities with technology as Demirobot or Demiagent. The concept of the demirobot is shown in Figure 3. Demirobot/demiagent has both aspect of human and robot agents. For example, when an avatar robot controlled by an operator performs customer services in a public space like a cafe, the avatar robot is regarded as an alter ego of the human operator, and as a robot that has different appearance from human at the same time. This means a demirobot is not just a human being, not just a robot/agent being, but mixture being (or quantum being) to expand human being.

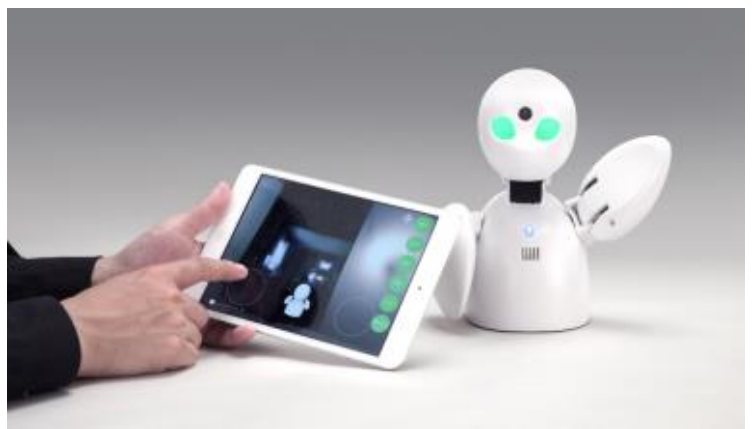

**Figure 2: Avatar Robot OriHime** [8].

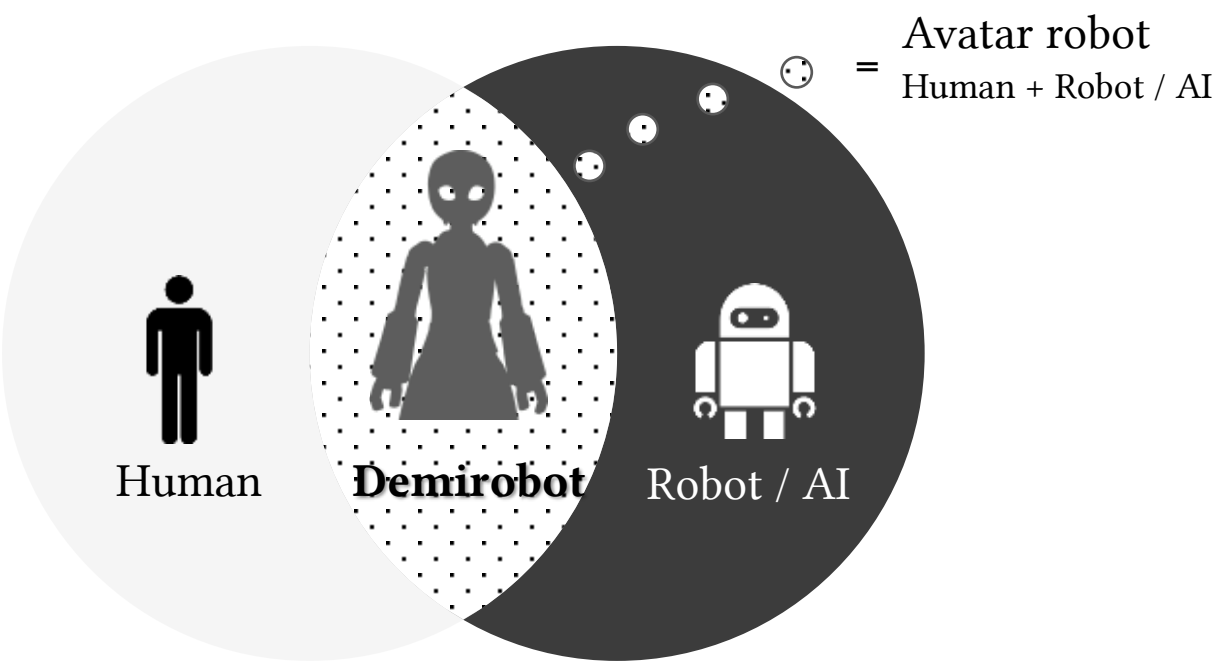

**Figure 3: Concept of Demirobot.**

The concept of demirobot is similar to the Wizard-of-Oz (WOZ) [9]. The difference between the concept of demirobot and WOZ is as follows: In the concept of demirobot, an operator of a robot is more than just an experimenter (wizard) that simulates system behavior, where the character/personality of each operator is important in the same manner as the human-human interaction. In the research of WOZ, human operator's behavior have been analyzed as interaction tactics, where the tactical factors have been suggested for human-robot agent interaction [10]. In contrast, a character/personality of each operator is important in the concept of demirobot. For instance, physical and facial expressions of personality for avatar robot have been proposed [11] [12], and these works are expected to be implemented to demirobot.

For designing a human-demirobot interaction, several interactions should be considered in the same manner as a human-robot interaction, e.g., a user interaction between human-robot system interfaces, a human-robot agent interaction, etc. In addition, a human-human social interaction should be considered. These designs of the human-demirobot interaction should be implemented and verified in stages because of a potential for challenges inherent in human-demirobot interaction.

In this study, as a first step to social implementation of avatar work, the telework that enables disabled people to engage in physical works, we develop an avatar robot OriHime-D, implement an avatar work by people with disabilities as an avatar robot cafe, and evaluate remote employment by people with disabilities using the OriHime-D.

## 3 DEVELOPMENT OF AVATAR ROBOT ORIHIME-D

In this study, we aim to realize social participation by those who cannot work because of their physical disabilities even if they want to. We propose an avatar robot OriHime-D, which enables disabled people who have difficulty going out to engage in physical work.

The OriHime-D has moving function and robotic hands in addition to communication functions of the OriHime. The OriHime-D enables an operator to operate interactive telework involving physical work such as carry things or serving customers remotely. This makes it possible to realize teleworks that requires moving the body, such as serving customers in cafe or restaurant, guidance in a building, giving instructions while looking around work sites, etc..

### 3.1 Appearance Design of OriHime-D

Figure 4 shows the appearance of the avatar robot OriHime-D. The size of OriHime-D is 500 [mm] in length, 400 [mm] in width, 1,180 [mm] in height, and 20 [kg] in weight. The maximum speed is 0.72 [km/h]. The battery driving time is about 6 hours.

Determine the size of the OriHime-D, we assume average height of a 6-year-old child, for many robots are built up to the size of children considered as being more friendly [13].

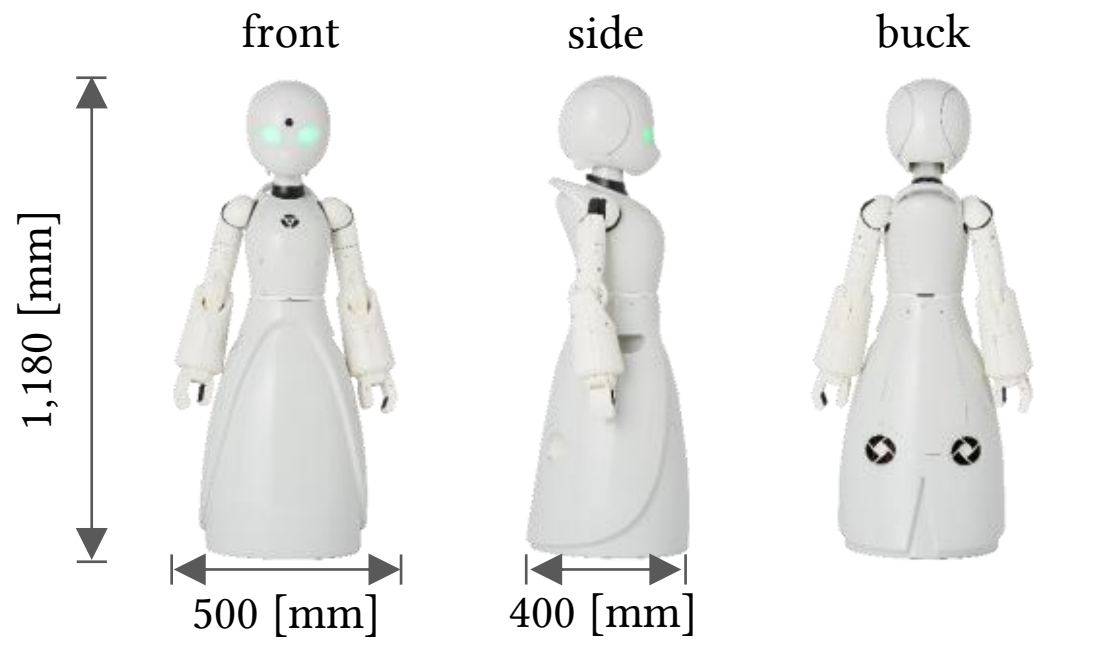

**Figure 4: Appearance of OriHime-D.**

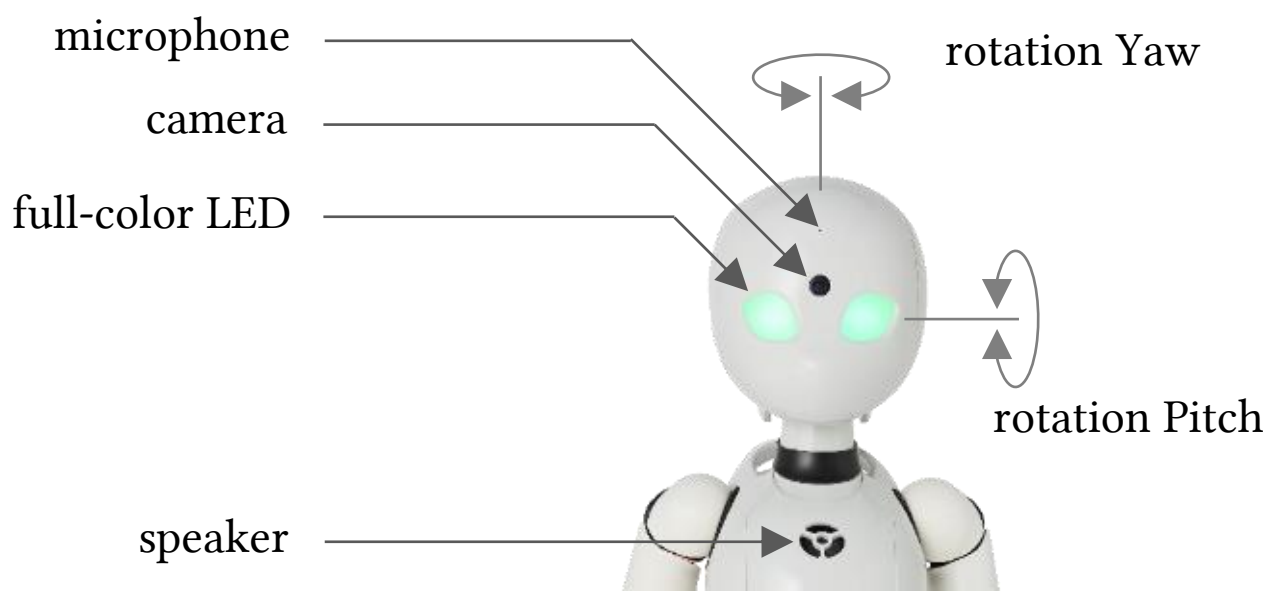

**Figure 5: Mechanism of the head part of OriHime-D.**

We design the OriHime-D's face in the motif of Noh mask, because it does not express personal characteristics but expresses human feature.

The head and body of the housing are made by cutting out resin. The arm is made of ABS using a 3D printer, for the arm need to be rapidly collected/modified depending on the intended use.

## 3.2 Mechanism of OriHime-D

*3.2.1 Head and chest*

Figure 5 shows the mechanism of the head of the OriHime-D. The neck has 2 degrees of freedom (D.O.F). The OriHime-D has full-color LEDs on the eye, a fisheye camera and microphone in the forehead, and a speaker on the chest.

An operator can control the head in order to get his/her surroundings by camera video, and to express nonverbal reaction for interlocutors using prepared motion sets. To get operator's surroundings the following four motions are prepared as: "look up", "look down", "look right", and "look left". To express reaction, the following three motion sets are prepared as: "one nod" for 'Yes', "shaking the head" for 'No', and "two nod" for supporting response. Thus, a total of seven motions is prepared for head.

*3.2.2 Arms*

Figure 6 shows the mechanism of the arm. Each arm has 6 D.O.F. An operator can control the arms for tasks to work on and nonverbal reaction in communication. To express reaction in communication, the following four motion sets are prepared as: "raise one hand", "bye-bye", 'hold up fists', "and "power pose". a total of seven motions is prepared for head.

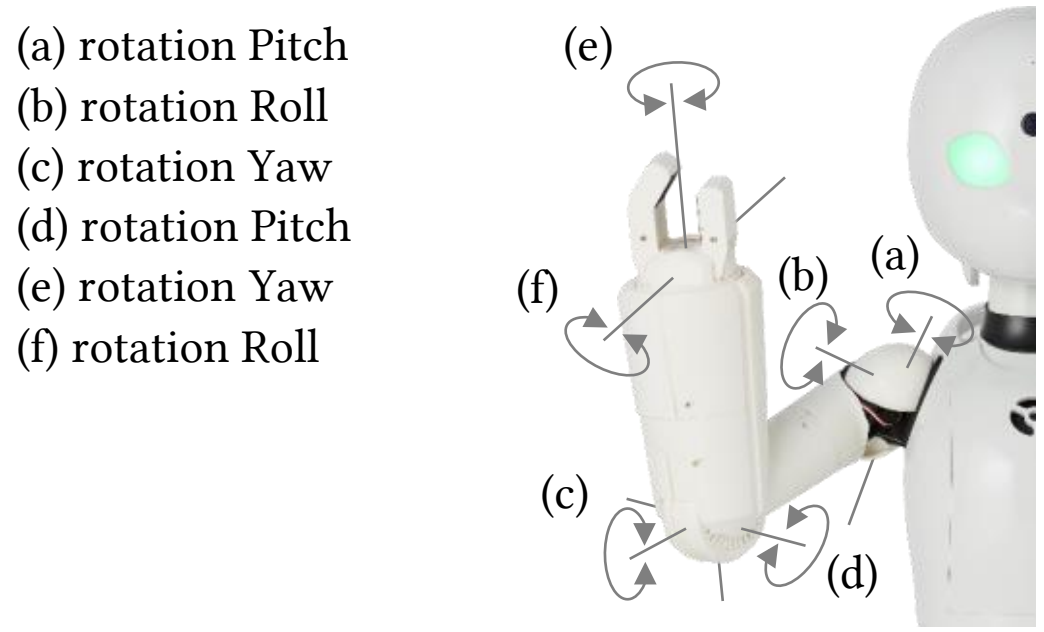

**Figure 6: Mechanism of the arm part of OriHime-D.**

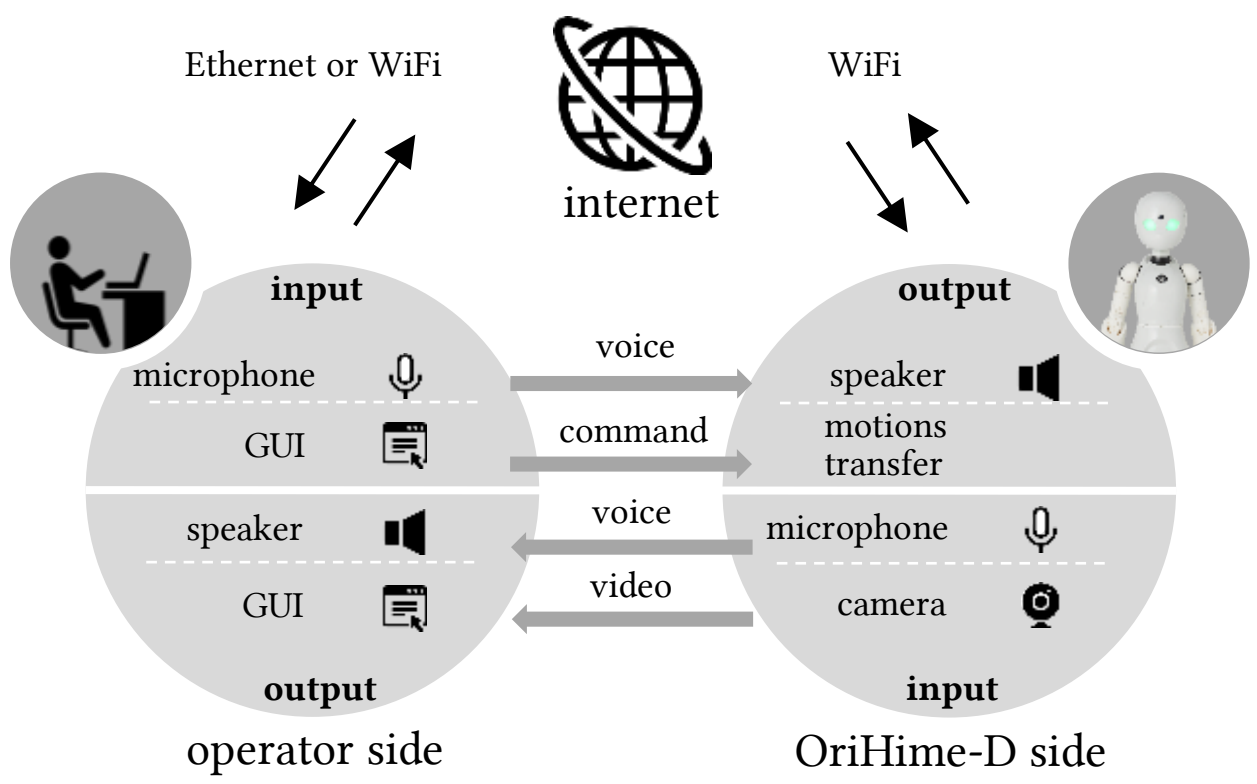

**Figure 7: System configuration diagram of OriHime-D.**

*3.2.3 Mobility*

The OriHime-D has two omni wheels with motors for locomotion, where the robot can go forward, go backward, turn left, and turn right. To reduce the workload of the operator, the OriHime-D has line tracing function for long distance movement, where the robot automatically moves to the preset target position.

## 3.3 System configuration of OriHime-D

The OriHime-D has two types of input method for its operating: a mouse operation and an operation using the line-of-sight input device OriHime eye (manufactured by Ory Laboratory Inc. [14]). Disabled people as operators can choose it depending on their own disabilities.

Figure 7 shows the system configuration of the OriHime-D. The operator remotely confirms his/her surrounding situation from the video transmitted from the camera on the OriHime-D's head. The voice of the interactors near the robot and the ambient sound are transmitted to the operator by the microphone on the forehead of the robot. The operator's voice is output from the speaker on the chest.

The graphical user interface is prepared for the operator, where the operator can communicate with the interlocutor by speech and selecting the prepared motions (seven motions for the head, four motions for the arms, and four for the locomotion). The specific appearance of the GUI is confidential.

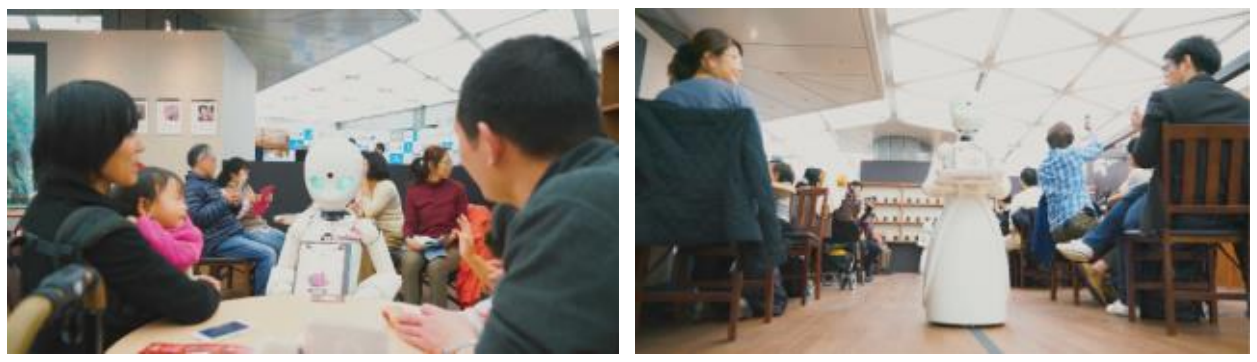

Figure 8: Scenes of the experimental cafe DAWN Ver. β.

Table 1: Pilot's disease name and computer operation.

| | disease name | operation method of laptop |
|---|---|---|
| 1 | | hand |
| 2 | ALS | eye |
| 3 | | eye |
| 4 | AVM | hand |
| 5 | Cervical Cord Injury | mouth |
| 6 | Myelitis | hand |
| 7 | SMA | hand and mouth |
| 8 | | hand |
| 9 | Somatoform Disorders | hand |
| 10 | Spinal Cord Injury | eye |

## 4 DEMONSTRATION EXPERIMENT ON AVATAR ROBOT CAFE

In order to clarify of the challenge of the proposed avatar work by disabled people using the avatar robot OriHime-D, we open up an avatar robot cafe "DAWN Ver.β" (hereinafter called experimental cafe).

### 4.1 AVATAR ROBOTS CAFE "DAWN Ver.β"

The experimental cafe have been held on the first floor of the Nippon Foundation Building of the Nippon Foundation in Minato-ku, Tokyo, for two weeks from November 26 to 30, 2018 and December 3 to 7, in a total of 10 days excluding Saturday and Sunday.

In the experimental cafe, the one session is total 60 minutes, where each costumer are served for 50 minutes according to a routine workflow. The workflow of the session is as follows: (i) opening talk (for 5 minutes by the cafe owner), (ii) order confirmation (for 10 minutes by the OriHime-D), (iii) drink serving (for 10 minutes by the OriHime-D), (iv) free talk with customers (for 20 minutes by the OriHime or the OriHime-D), (v) ending talk (for 5 minutes by the cafe owner). With that total 50 minutes, and after 5 minutes break, customers of next session are coming in the cafe (for 5 minutes). As a results, the pilots have 55 minutes actual working time per one hour business time.

We hold four session per day, that is, the experimental cafe runs four hours in total per day (except three sessions/three hours on November 26).

Figure 8 shows the scenes of experimental cafe, where three OriHime-D and two OriHime work all the time. There are six tables in the cafe, and four customers from one to are sitting at each table. The OriHime-D can move around the cafe by themselves. The OriHime is setting on the table, and move between the tables with support of the human staffs.

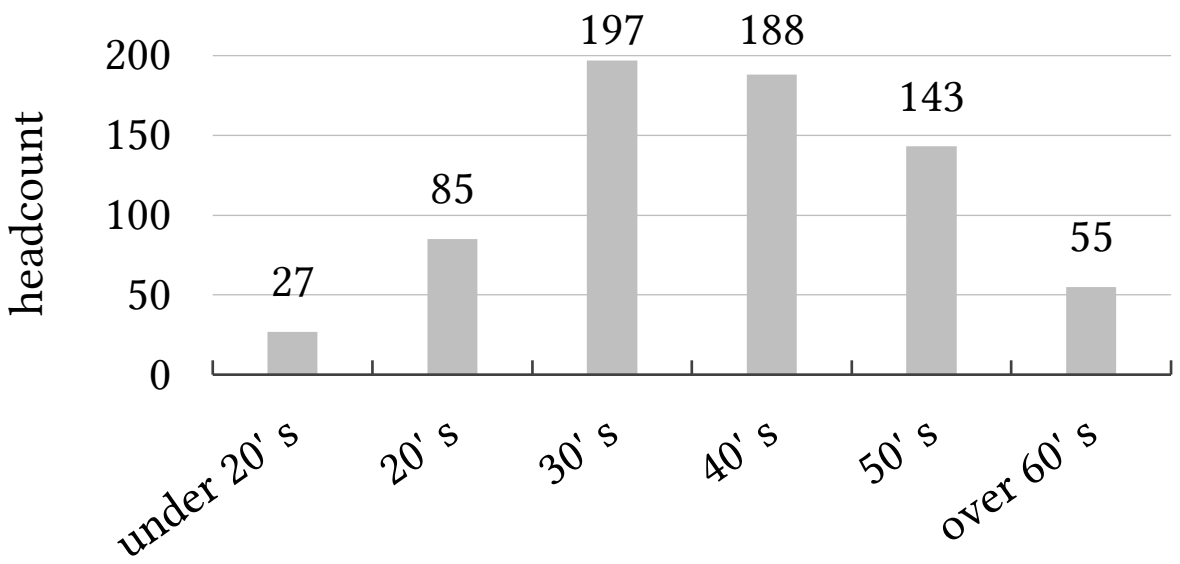

Figure 9: The number of customers by age.

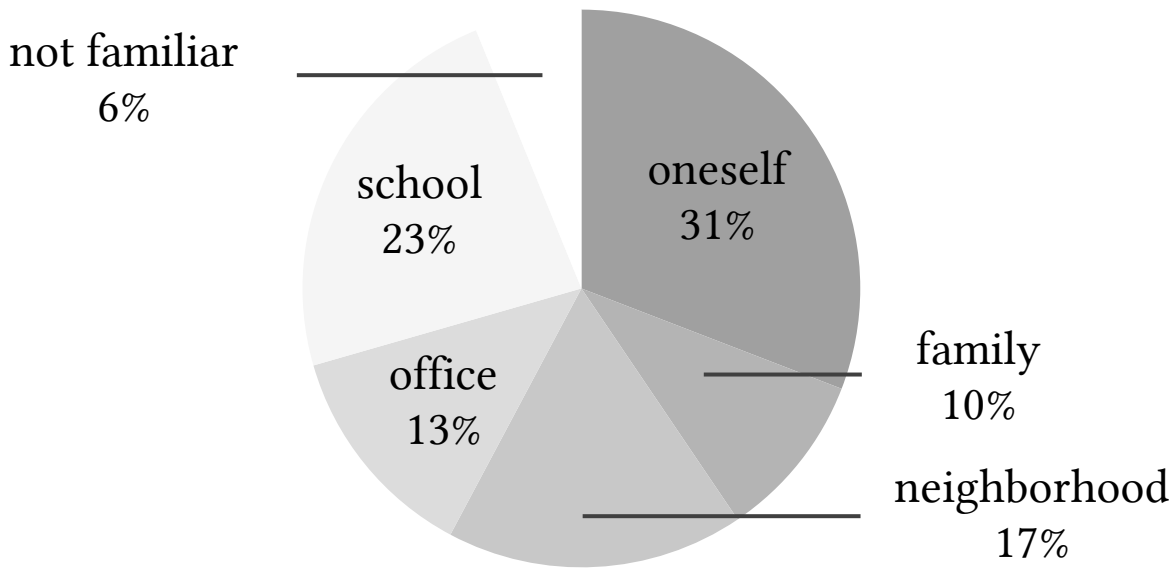

Figure 10: People with disabilities who is close to visitors.

### 4.2 DISABLED PEOPLE OPETATING ABATAR ROBOT

In the experimental cafe, the operators of the OriHime-D are five males and five females, a total of 10 disabled people (hereinafter called pilots) including patients of ALS (Amyotrophic Lateral Sclerosis), SMA (Spinal Muscular Atrophy), AVM (Autophagic Vacuolar Myopathy), Spinal Cord Injury, Cervical Cord Injury, Myelitis, and Somatoform Disorders, and perform avatar work.

Some pilots are designated as having difficulty going out because of severe disabilities or designated intractable diseases. Three pilots operate his/her personal computer using the gaze input device, five pilots operate using mouse with hands, and two operate using mouse with jaw. Table 1 gives the classification for the pilot disease and the PC operation methods.

In this experimental cafe, we assign work shifts to the pilots. The shifts need to be flexible considering their physical condition for the safety of pilots. Thus, the number of the day of working and the business hour differ in individuals.

### 4.3 CUSTOMER OF EXPERIMENTAL CAFE

Figure 9 shows the customer traffic by age who cooperated in the questionnaire at the experimental cafe (hereinafter this customers are called the questionees). Figure 10 shows the existence or non-existence of disabled people close to the questionees (multiple answers allowed). There were a total of 700 questionees including 335 men, 348 women, 3 others, and 14 unanswered. In addition, the number of the direct media coverage of the experimental cafe is a total of 55 including 45 in Japan, 1 in Korea, 2 in the United States, 1 in Russia, 1 in Germany, 1 in France, 2 in Spain, and 2 in the UK.

## 5 EVALUATION OF AVATAR WORK BY DISABILITY PEOPLE USING AVATAR ROBOT ORIHIME-D

In order to disseminate avatar work by people with disabilities, it is necessary for pilots to continuously and aspiringly engage in the avatar work depending on their disabilities. In 5.1, we evaluate the effects of the proposed avatar work in experimental cafe using the proposed the OriHime-D on the pilots with disabilities. In 5.2, we focus on the change of psychophysical and physical conditions of the pilots of ALS (Amyotrophic Lateral Sclerosis) patients, for they especially have a large gap between their own body and the avatar robot. In 5.3, the results of staff attitude surveys for avatar work and employment by the pilots with disabilities are described.

**Table 2: Face scale for psychological and physical conditions.**

| | score | | | | | | | | | |
|---|---|---|---|---|---|---|---|---|---|---|
| | negative | | | | | | | | positive | |
| | 1 | 2 | 3 | 4 | 5 | 6 | 7 | 8 | 9 | 10 |
| mood | anger ⟷ relax | | | | | | | | | |
| fatigue | fatigue ⟷ relax | | | | | | | | | |
| fullness | sad ⟷ happy | | | | | | | | | |

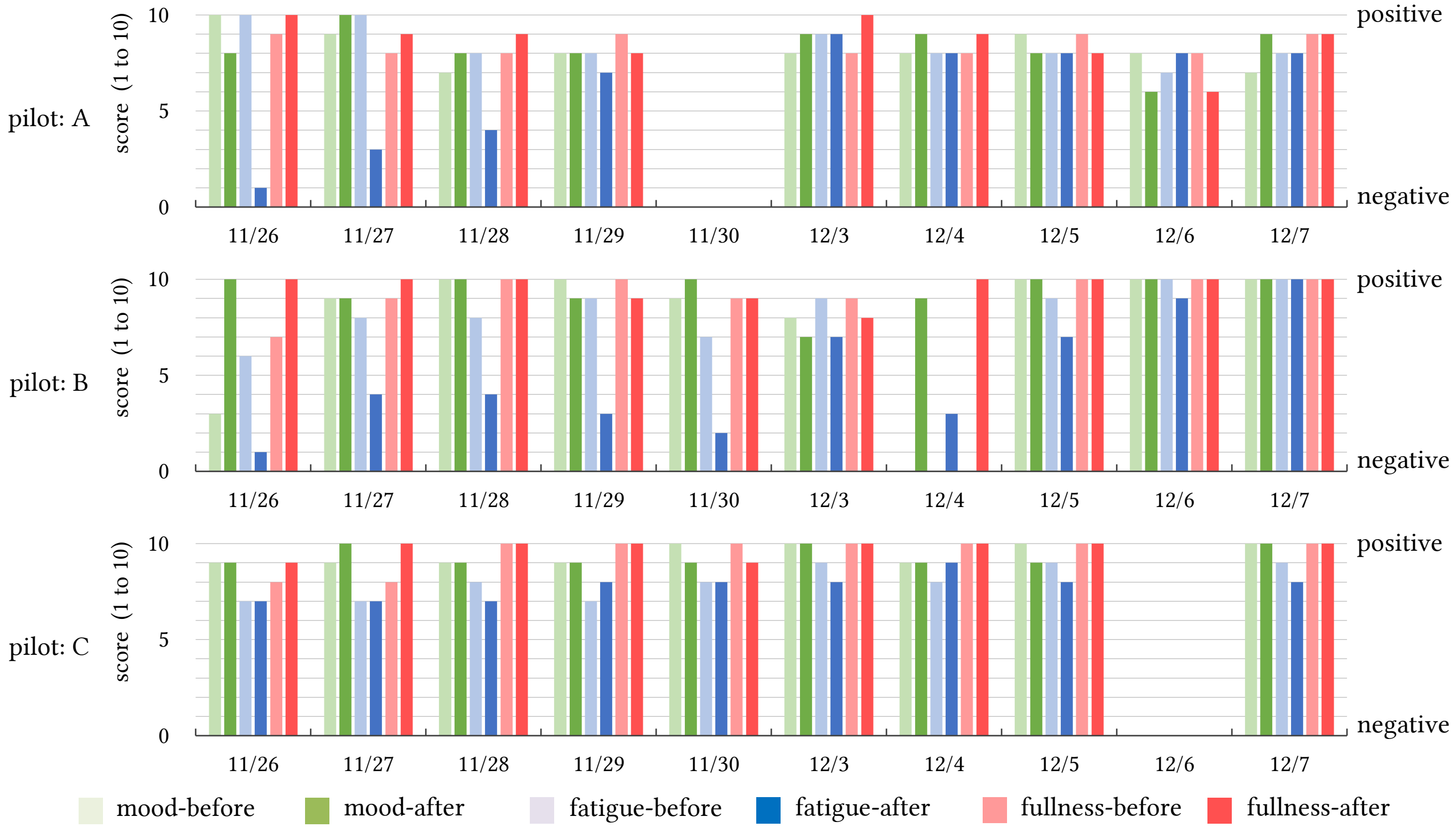

**Figure 11: Feelings, fatigue and fulfillment of the pilots before and after the avatar work.**

**Table 3: The number of sessions the plots work in (equal to working hours). In the session, the pilot is allotted the OriHime-D or the OriHime.**

| pilot | robot | working time [h] 11/26 Mon | 11/27 Tue | 11/28 Wed | 11/29 Thu | 11/30 Fri | 12/3 Mon | 12/4 Tue | 12/5 Wed | 12/6 Thu | 12/7 Fri |
|---|---|---|---|---|---|---|---|---|---|---|---|
| A | OriHime | 1 | 2 | 1 | 1 | 0 | 2 | 0 | 2 | 2 | 0 |
| | OriHime-D | 2 | 2 | 2 | 0 | 0 | 2 | 2 | 1 | 0 | 2 |
| B | OriHime | 0 | 1 | 0 | 1 | 1 | 2 | 1 | 2 | 1 | 0 |
| | OriHime-D | 3 | 3 | 3 | 2 | 1 | 2 | 2 | 0 | 2 | 1 |
| C | OriHime | 3 | 4 | 1 | 3 | 1 | 1 | 0 | 1 | 0 | 1 |
| | OriHime-D | 0 | 0 | 3 | 1 | 3 | 2 | 2 | 2 | 0 | 3 |

### 5.1 EFFECTS FOR PSYCHOLOGICAL AND PHYSICAL CONDITIONS BY AVATAR WORK

We evaluate the effects of psychological and physical conditions for the pilots engaging in the avatar work using the proposed robots the OriHime-D and OriHime. Three types of face scales are used to investigate the effects of psychological and physical conditions for mood, fatigue, and fullness. Each face scale has 10 scales for relax-anger (mood), relax-fatigue (fatigue), and happy-sad (fullness) respectively. Pilots use Google Forms to answer the questionnaires. Table 2 gives the face scale used in the evaluation.

*5.1.1 Experimental procedure*

Step 1) The pilot answers the 3 types of face scales before work.

Step 2) The pilot answers the 3 types of face scales after work.

*5.1.2 Experimental result*

We focus on the results of three pilots with the longest working hour among the 10 pilots because of space limitations. Figure 11 shows the results of converting the face scale score according to Table 3 gives the number of sessions (equal to working hours) that the plots work in of the three pilots. In Figure 11, the pilot B did not answer at pre-work. We summarize the number of increase and decrease of each evaluation score before and after the work in Table 4.

*5.1.3 Discussion*

We compare the scores before and after work for each item separately.

As Table 4 given, we compare the number of days of mood change before and after work. During a total of 29 business days, the feeling after work increased for 8 days compared to before work, and the feeling fell for 6 days. This indicates the avatar work affects mood. Depending on the work design, avatar work has the potential to lead to finding motivation in life for people with disabilities. Also, from Figure 11, it was only one day that the mood changed by more than three levels comparing before and after the work (Pilot B on 11/26, $26^{th}$ Nov.).

Regarding fatigue, it can be confirmed that pilot A and pilot B have less fatigue after work. Regarding Pilot C, no trend related to fatigue could be confirmed.

### 5.2 CHANGES OF PSYCHOLOGICAL AND PYSICAL CONDITIONS OF ALS PATIENTS AT AVATAR WORK

We focus on the change of psychophysical and physical conditions of the pilots of ALS (Amyotrophic Lateral Sclerosis) patients, for they especially have a large gap between their own body and the avatar robot OriHime-D. For detailed exploration, we measure smile expression time of the pilot during the work in addition to the face scale described in 5.1. The ALS pilot operates the OriHime-D with synthesized speech by gaze input. For the measurement, videos are taken from the front and the back of the ALS pilot with cameras. Figure 12 shows the scene of the avatar work by the ALS pilot.

*5.2.1 Experimental procedure*

Step 1) The experimental collaborator at the ALS pilot's home start to shoot a video before the work.

Step 2) The ALS pilot work according to the shift schedule including the face scale investigation described in 5.1.

Step 3) After the work, the experiment cooperator stops taking the video.

Step 4) Perform steps 1) to 3) on the working day.

*5.2.2 Experimental result*

The results of the face scale investigation by the ASL pilot is shown in Figure 13, where the scores are converted according to Table 2. Table 5 gives the ALS pilot's actual work time (seconds), smile expression time (%), customer service time (%), and the number of customers. Figure 14 shows the ratio of smile time per working hours. One session has one hour of working hour including 3300 seconds (55 minutes) of actual working time and 300 seconds (5 minutes) of short break. Figure 15 shows the contents breakdown of the actual working time.

*5.2.3 Discussion*

First we focus on the score of fatigue to examine the possibility of continuous avatar work by ASL pilot, where the scores of the after work on 11/29, 12/3, 12/4, and 12/6 have shown increases compared to that of before work. The difference in fatigue scores between before and after work is the highest on the first day, 11/29. This indicates the passivity that the ASL pilot adapts to the avatar work, and this reduce fatigue as a result. Only on the last day of the work (12/7), the ASL pilot have recovered from fatigue after work. We think this is due to the effect of the adaptation to avatar work, in addition to the actual work time (3300 [s]) that was half of that of all the other days and included the smallest active time (service + movement). Since the score of fulfillment level was lowered after work on the same day, the ALS pilots may have been felt that the work volume was unsatisfactory.

**Table 4: Number of rises and falls in evaluation values before and after the avatar work.**

| pilot | | A | B | C | Total |
|---|---|---|---|---|---|
| Day of duty | | 9 | 10 | 9 | 29 |
| mood | positive | 5 | 2 | 1 | 8 |
| | negative | 2 | 2 | 2 | 6 |
| fatigue | positive | 1 | 0 | 2 | 3 |
| | negative | 4 | 8 | 4 | 16 |
| fulfilling | positive | 5 | 2 | 2 | 9 |
| | negative | 2 | 2 | 1 | 5 |

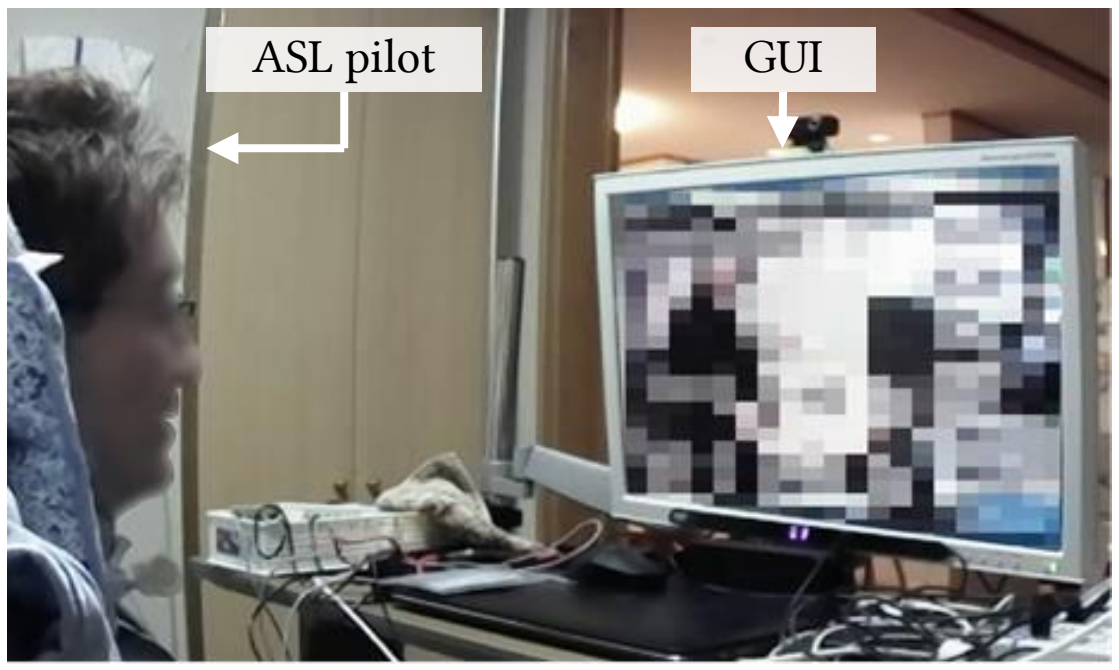

**Figure 12: The Avatar Work sense of the ALS pilot.**

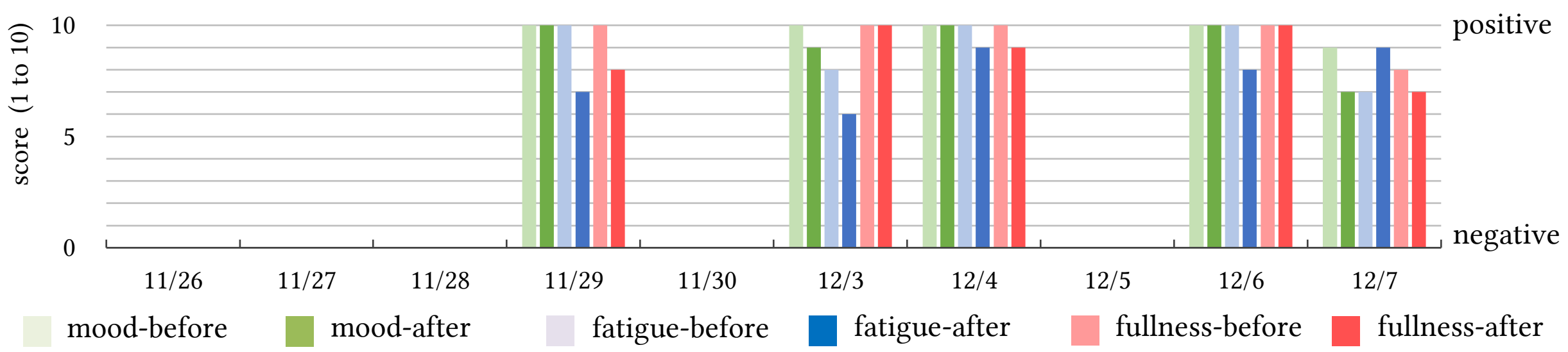

**Figure 13: ALS Pilot's mood, fatigue, and fulfillment before and after the avatar work.**

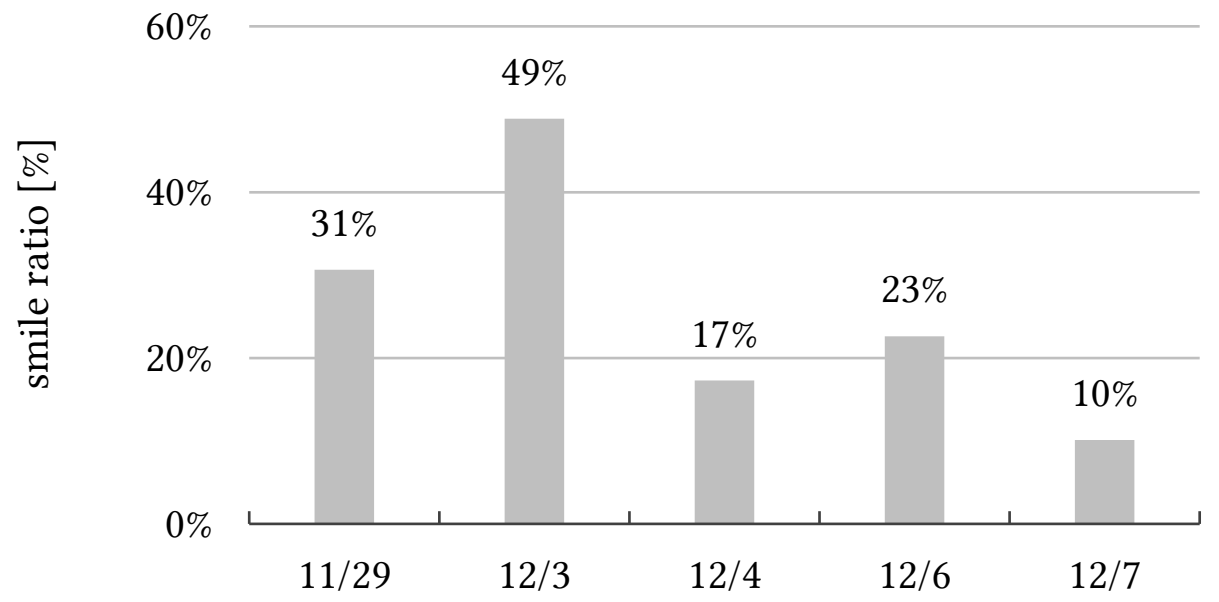

**Figure 14: Ratio of smile expression time per working hours.**

**Table 5: Actual working time, smile expression time, customer service time, and number of customers.**

| | 11/29 Thu | 12/3 Mon | 12/4 Tue | 12/6 Thu | 12/7 Fri |
|---|---|---|---|---|---|
| working time [s] | 6600 | 6600 | 6600 | 6600 | 3300 |
| smile time rate [%] | 31 | 49 | 17 | 23 | 10 |
| customer service time [%] | 25 | 32 | 33 | 17 | 18 |
| No. of customers | 21 | 38 | 21 | 24 | 19 |

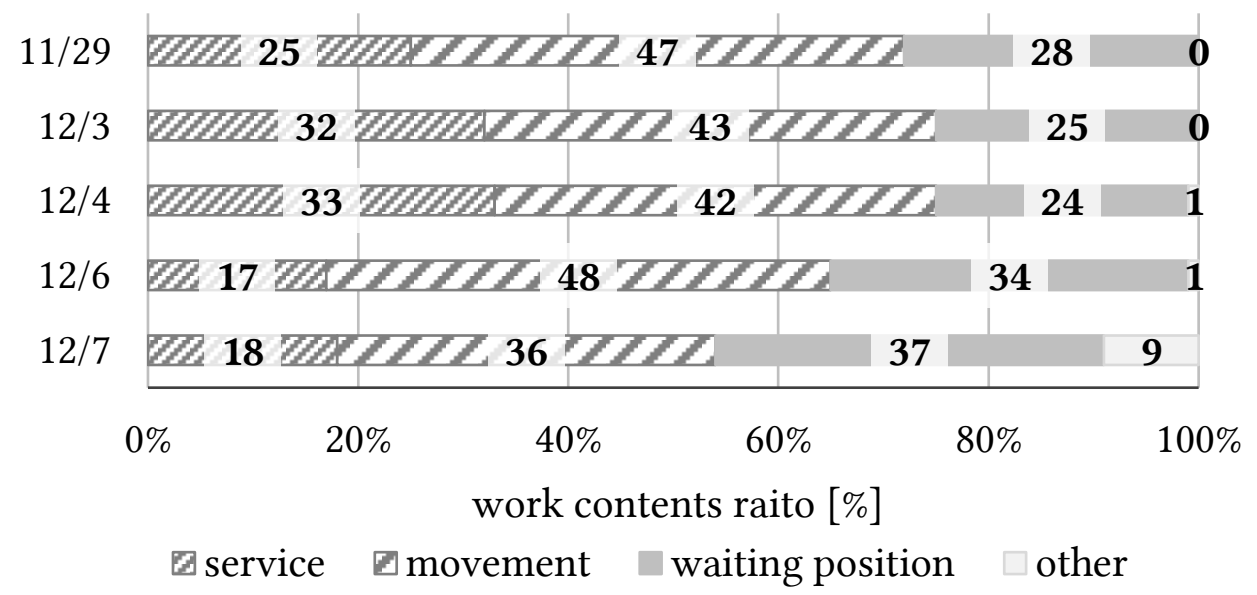

**Figure 15: Breakdown of work contents. Each ratio shows the percentage of the performing time per actual working time.**

Second we consider smile expression time. We focus on 12/3, which shows the significantly higher rate of smile expression time during work hours. This day includes the highest time of the active time (service + movement) per the actual working time (see Figure 15). At the same time the number of the customers who the ASL pilot serve is highest (see Table 5). As results of the face scales, the fulfilling score is kept high (ten) before and after work, though the fatigue score is lowest after work (see Figure 13). From this we infer that the avatar work of customer service have led to mental fulfillment as a result, although the pilot has had declines in fatigue and mood due to customer service.

For the reasons stated above, we conclude that it is possible to design avatar work leading to mental fulfillment with adaptable workload.

## 5.3 STAFF ATTIUDE SURVEY FOR AVATAR WORK AND EMPLOYMENT BY PEOPLE WITH DISABILITIES

We conduct staff attitude staff attitude surveys for avatar work and employment to the pilots with disabilities.

*5.3.1 Experiment*

After the implementation period of the experimental cafe, we have conducted the questionnaire surveys about his/her rewarding for the avatar work and social participation. The pilots evaluate five items on a scale of one to five respectively. The items are as follows: (i) rewarding, (ii) self-fulfillment, (iii) aptitude for the work, (iv) actual feeling of social participation, and (v) appropriateness of the experimental cafe as an avatar work. The questions presented the pilots are as follows: (i) "Did you feel your work in the experimental cafe rewarding?", (ii) "Do you think you have achieved self-fulfillment through this work experience?", (iii) "Do you think you could utilize your capability through this work?", (iv) "Do you think you could be of service to people and society through this work?", and (v) "Do you want to intentionally invite new people who you don't know to work at the avatar robot cafe next time around? ".

Prepare a total of 5 items, and evaluate them on a scale of 1 to 5, respectively.

The 9 of 10 pilots have answered the questionnaire within a week (one is unanswered).

*5.3.2 Experimental result*

Figure 16 shows the average value of the questionnaire results of 9 pilots. All items show a rating of 4 or higher.

*5.3.3 Discussion*

The fifth item, (v) appropriateness of the experimental cafe as an avatar work have shown an average of 4.9, the highest evaluation value. At the same time, the fourth item, (iv) actual feeling of social participation have shown an average of 4.7. From these results, we conclude the work content of the experimental cafe is appropriate for people with a variety of disabilities seeking social participation as an avatar work for the first time. In these two items, all questionees have answered 4 or more.

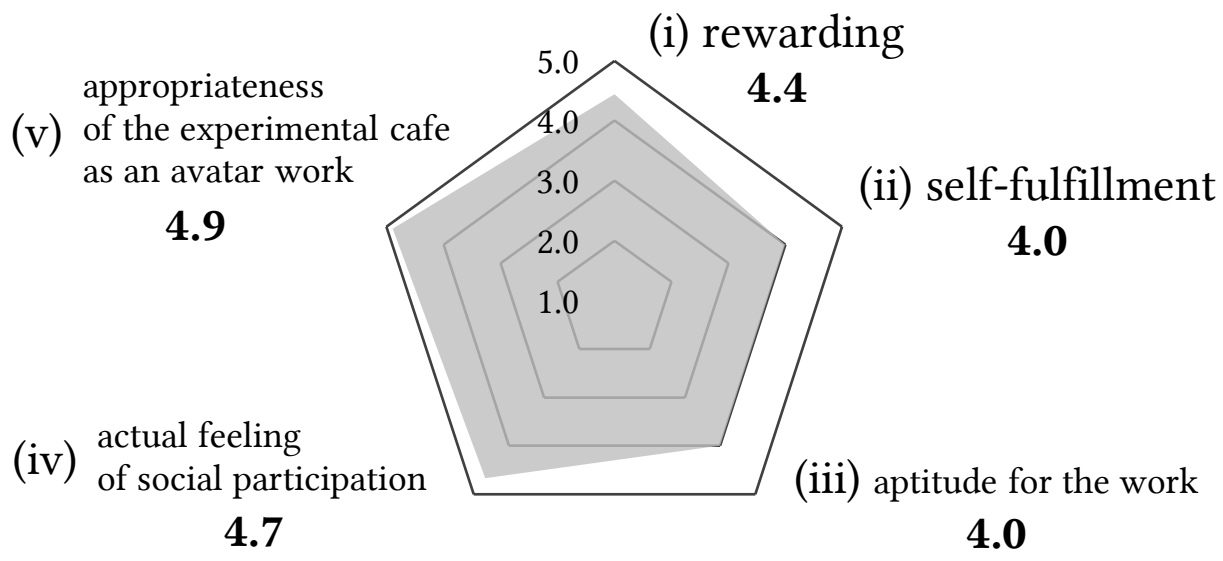

**Figure 16: Ratio of smile expression time during work.**

Regarding (i) rewarding and (ii) self-fulfillment, there were questionees who have answered an evaluation value of 2, while showing an average value of 4.0 or more (one questionee for each). From this, we think that it is necessary for people with disabilities to choose a work from various avatar work depending on their abilities as the next step of this experimental cafe. In addition, we intend to evaluate the quality of service for the next step in order to realize a wide range of avatar works.

## 6 CONCLUSION

In this study, we have developed the avatar robot OriHime-D, which people with disabilities can operate depending on their disabilities, and have proposed avatar work as telework that enable to engage in physical works such as customer service. As the results of two weeks of experiment in the avatar robot cafe by 10 people with disabilities, we have confirmed that the proposed avatar work leads to mental fulfillment for people with disparities, and can be designed with adaptable workload. As the results of the staff attitude staff attitude surveys for 9 of 10 pilots, we have confirmed that the work content of the experimental cafe is appropriate for people with a variety of disabilities seeking social participation. In addition, the need for choices of avatar work has been suggested.

To realize a wide range of avatar works, various semi-automated behaviors of the avatar robot are required, and we are addressing this as a further challenge.

In addition, legal issues are confirmed as future issues for the social implementation of this research. Severely disabled people who require constant care receive public assistance for care costs, but there is a problem, where they cannot receive this assistance during their economic activities. The solution to this problem is indispensable for the real social implementation of avatar work for people with disabilities in the true sense, and now we are working with governments such as Kanagawa Prefecture.

With the spread of avatar robots and avatar work, it is possible to realize a society where we can do anything if we have free mind, even though we are bedridden. This contributes to fulfillment all through life and lifetime working, and at the same time leads to a solution to the employment shortage problem.

This study was approved by the Ethical Review Board for the use of human subjects of Kanagawa Institute of Technology (No.20181122-02).